\documentclass[runningheads]{llncs}
\usepackage{graphicx}
\usepackage{amsmath,amssymb} 
\usepackage{color}
\usepackage{graphicx}
\usepackage{amsmath,amssymb} 
\usepackage{color}
\usepackage{threeparttable}
\usepackage[linesnumbered, ruled]{algorithm2e}
\SetKwRepeat{Do}{do}{while}
\usepackage{graphicx}
\usepackage{amsmath}
\DontPrintSemicolon
\usepackage[noend]{algpseudocode}
\usepackage{amsmath,amssymb} 
\usepackage{color}
\usepackage{amsmath}
\DeclareMathOperator*{\argmax}{argmax}
\usepackage{hyperref}
\hypersetup{
  colorlinks   = true, 
  urlcolor     = black, 
  linkcolor    = red, 
  citecolor = blue 
}
\newcommand{\cmt}[1]{}
\begin{document}
\pagestyle{headings}
\mainmatter

\def\ACCV18SubNumber{756}  

\title{ Visual Re-ranking with Natural Language Understanding for Text Spotting}
 
\author{Ahmed Sabir\inst{1}, Francesc Moreno-Noguer\inst{2} and Llu\'{\i}s Padr\'o\inst{1}}

\institute{TALP Research Center, Universitat Polit\`ecnica de Catalunya, Barcelona, Spain \and
Institut de Rob\`otica i Inform\`atica Industrial (CSIC-UPC), Barcelona, Spain
\email{asabir@cs.upc.edu, fmoreno@iri.upc.edu, padro@cs.upc.edu}}
\maketitle

\begin{abstract}
Many scene text recognition  approaches are based on purely visual information and ignore the semantic relation between scene and text. In this paper, we tackle this problem from natural language processing perspective to fill the gap between language and vision. We propose a post-processing approach to improve scene text recognition accuracy by using occurrence probabilities of words (unigram language model), and the semantic correlation between scene and text. For this, we initially rely on an off-the-shelf deep neural network, already trained with large amount of data, which provides a series of text hypotheses per input image. These hypotheses are then re-ranked using word frequencies and semantic relatedness with objects or scenes in the image. As a result of this combination, the performance of the original network is  boosted with almost no additional cost. We validate our approach on ICDAR'17  dataset.   
\end{abstract}

\section{Introduction}

\label{sec:intro}
Machine reading has shown a remarkable progress in  Optical Character Recognition systems (OCR). However, the success of most OCR systems is restricted to simple-background and properly aligned documents, while text in many real images  is affected by a number of artifacts including  partial occlusion, distorted perspective and complex backgrounds. In short, developing OCR systems able to read text in the wild is still an open problem. In the computer vision community, this problem is known  as \textit{Text Spotting}. However, while state-of-the-art computer vision algorithms have shown remarkable results in   recognizing object instances in these images, understanding and recognizing the included text in a robust manner is far from being considered a solved problem.  

Text spotting pipelines address the end-to-end problem of  detecting and recognizing text in unrestricted images (traffic signs, advertisements, brands in clothing, etc.). The problem is usually split in two phases: 1) \textit{text detection stage}, to estimate the bounding box around the candidate word in the image and 2) \textit{text recognition stage}, to identify the text inside the bounding boxes. In this work we focus on the second stage, an introduce a simple but efficient  post-processing approach based on Natural Language Processing (NLP) techniques. 

There exist two main approaches to perform text recognition in the wild. First, lexicon-based methods, where the system learns to recognize words in a pre-defined dictionary. Second, lexicon-free, unconstrained recognition methods, that aim at predicting character sequences.

In this paper, we propose an approach that intends to fill the gap between language and vision for the scene text recognition problem. Most recent state-of-the-art works focus on automatically detecting and recognizing text in unrestricted images from a purely computer vision perspective. In this work, we tackle the same problem but also leveraging on NLP techniques. Our approach seeks to integrate prior information to the text spotting pipeline. This prior information biases the initial ranking of candidate words suggested by a deep neural network, either lexicon based or not. The final re-ranking is based on the word frequency and on the semantic relatedness between the candidate words and the information in the image.

\begin{figure}[t!]
 \centering 
\includegraphics[width=3.7in]{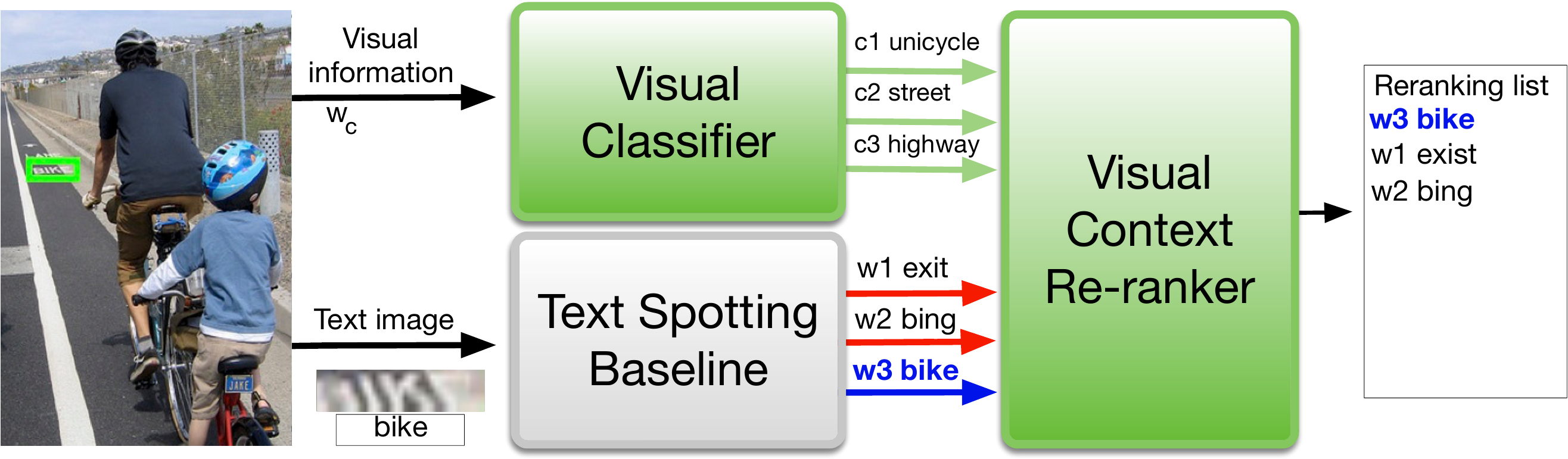}  
 \caption{Overview of the system pipeline. A re-ranking post-process using visual context information to re-rank the potential candidate word based on the semantic relatedness with the context in the image (where the text is located). In the example of the figure, the word \textit{bike} has been re-ranked thanks to the detected visuals (w$_c$) \textit{unicycle, street, highway}. }  
 \label{fig:quarter} 
 \end{figure}

Figure \ref{fig:quarter} shows an example where the candidate word \textit{bike} is re-ranked thanks to the visual context information \textit{unicycle, street, highway} detected by a visual classifier. This is a clear example that illustrates the main idea of our approach.          

Our main contributions include several post-processing methods based on NLP techniques such as word frequencies and semantic relatedness which are typically exploited  in NLP problems but less common in computer vision ones. We show that by introducing a candidate re-ranker based on word frequencies and semantic distance between candidate words and objects in the image, the performance of an off-the-shelf deep neural network can be improved without the need to perform additional training or tuning.
In addition, thanks to the inclusion of the unigram probabilities, we overcome the baseline limitation of false detection of short words of \cite{Max:16,Suman:17}. 

The rest of the paper is organized as follows: Sections \ref{sec:relatedwork} and \ref{sec:baseline} describe related work and our proposed pipeline. Sections \ref{sec:reranking} and \ref{sec:combining} introduce the external prior knowledge we use, and how it is combined. Section \ref{sec:experiments} presents experimental validation of our approach on a publicly available standard dataset. Finally, Sections \ref{sec:discussion} and \ref{sec:conclusion} summarize the result and specifies future  work.

\section{Related Work}
\label{sec:relatedwork}

\textit{Text spotting} (or end-to-end text recognition),  refers to the problem of automatically detecting and recognizing text in images in the wild. Text spotting may be tackled by either a  lexicon-based or a lexicon-free perspective. Lexicon-based recognition methods use a pre-defined dictionary as  a reference to guide the recognition. Lexicon-free methods (or unconstrained recognition techniques),   predict   character sequences without relying on any dictionary. The first lexicon-free text spotting system  was proposed by \cite{Lukas:10}. The system extracted  character candidates via maximally stable extremal regions (MSER) and eliminated non-textual ones  through a trained classifier. The remaining candidates were fed into a character recognition module,  trained using a large amount of synthetic data. More recently, several  deep learning alternatives have been proposed. For instance, PhotoOCR  \cite{Alessandro:13} uses a Deep Neural Network (DNN) that performs end-to-end text spotting using    histograms of oriented gradients  as input of the network. It is a lexicon-free  system able to read characters in uncontrolled conditions. The final word re-ranking is performed by means of two language models, namely a character and an $N$-gram language model. This approach combined two language models, a character based bi-gram model with compact 8-gram and 4-gram word-level model. Another approach employed language model for final word re-ranking \cite{Anand:12}. The top-down integration can tolerate the error in text detection or mis-recognition. 

Another DNN based approach is introduced by  \cite{Max:16}, which applies a sliding window over Convolutional Neural Network (CNN) features that use a fixed-lexicon based dictionary. This is further extended in \cite{Max:14}, through a deep architecture that allows feature sharing. In \cite{Baoguang:16} the problem is addressed using a Recurrent CNN, a novel lexicon-free neural network architecture  that integrates  Convolutional and Recurrent Neural Networks for image based sequence recognition. Another sequence recognition approach \cite{Suman:17} that uses LSTM with visual attention mechanism for character prediction. Although this method is lexicon-free, it includes a language model to improve the accuracy. 
Finally most recently, \cite{Yunze:17} introduced a CNN with connectionist temporal classification (CTC) \cite{Alex:06} to generate the final label sequence without a sequence  model such as LSTM. This approach use stacked convolutional to capture the dependencies of the input sequence. This algorithm can be integrated with either lexicon-based or lexicon-free recognition. 

\begin{figure*}[t!]
\centering 
\includegraphics[width=0.8\textwidth]{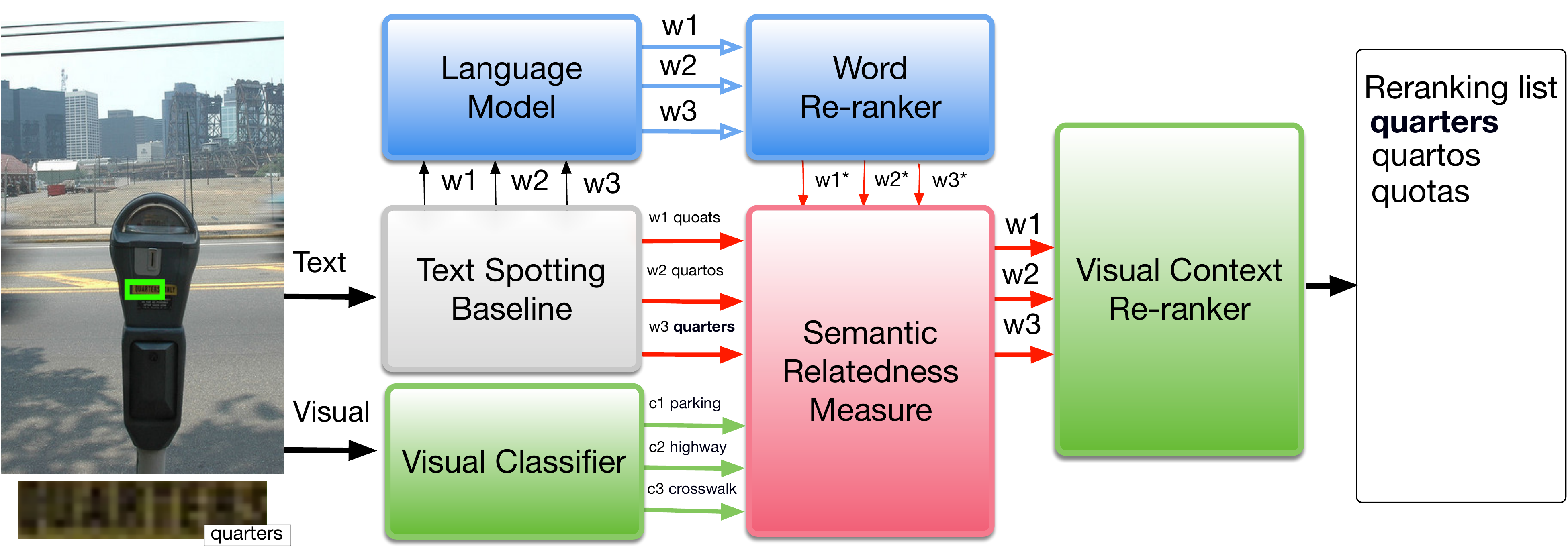}
\caption{Scheme of the proposed visual context information pipeline integration into the text spotting system. Our approach uses the language model and a semantic relatedness measure to re-rank the word hypothesis. The re-ranked word \textit{quarters} is semantically related with the top ranked visual \textit{parking}. See more examples in Figure \ref{fig:moreexamples}.}
\label{fig:happy} 
\end{figure*}

However, deep learning methods --either lexicon-based or lexicon-free-- have drawbacks: Lexicon-based approaches need a large dictionary to perform the final recognition. Thus, their accuracy will depend on the quality and coverage of this lexicon, which makes this approach unpractical for real world applications where the domain may be different to that the system was trained on. On the other hand, lexicon-free recognition methods rely on sequence models to predict character sequences, and thus they may generate likely sentences that do not correspond to actual language words. In both cases, these techniques rely on the availability of large datasets to train and validate, which may not be always available for the target domain.

The work of \cite{Yash:16} also uses  visual prior information to improve the text spotting task, through a new lexicon built with Latent Dirichlet Allocation (LDA) \cite{David:03}. The topic modeling learns the relation between text and images. However, this approach relies on captions describing the images rather than using the main key words semantically related to the images to generate the lexicon re-ranking. Thus, the lexicon generation can be inaccurate in some cases due to the short length of captions. In this work, we consider a direct semantic relation between scene text and its visual information. Also, unlike \cite{Yash:16} that only uses visual information over word frequency count to re-rank the  most probable word, our approach combines both methods by leveraging also on a frequency count based language model.

\section{General  Description of our Approach}
\label{sec:baseline}

Text recognition approaches can be divided in two categories: (a) character based methods that rely on a single character classifier plus some kind of sequence modeling (e.g. n-gram models or LSTMs), and (b) lexicon-based approaches that intend to classify the image as a whole word.

In both cases, the system can be configured to predict the $k$ most likely words given the input image. Our approach focuses on re-ranking that list using language information such as word frequencies, or semantic relatedness with objects in the image (or \texttt{visual context}) in which the text was located.
\subsection{Baseline Systems}
\label{sec:baselinesystems}
We used two different  off-the-shelf baseline models: First, a CNN \cite{Max:16} with fixed lexicon based recognition. It uses a fixed dictionary containing around 90K word forms. Second, we considered a LSTM architecture with a visual attention model \cite{Suman:17}. The LSTM generates the final output words as character sequences, without relying on any lexicon. Both models are trained on a synthetic dataset \cite{Jaderberg14c}. The output of both models is a vector of softmax probabilities for candidate words.
For each text image the baseline provides a series of $k$ text hypotheses, that is fed to our model. Let us denote the baseline probability of any of the $k$ most likely words ($w_j, 1\le j\le k$) produced by the baseline as follows: 
\begin{equation} 
\small \begin{array}{l}P_{BL}(w_j)\;=\;softmax(w_j,BL)\\\\\end{array}
\end{equation}

\subsection{Object Classifier}
Next, we will use out-of-the-box state-of-the-art visual object classifiers to extract the image context information that will be used to re-rank candidate words according to their semantic relatedness with the context.

We considered three pre-trained CNN classifiers: ResNet \cite{Kaiming:16}, GoogLeNet \cite{Christian:15} and Inception-Resnet-v2 \cite{szegedy2017inception}. The output of these classifiers is a $1000$-dimensional vector with the probabilities of $1000$ object instances. In this work we consider a threshold of most likely objects of the context predicted by the classifier. Additionally, we use a threshold to filter out the probability predictions when the object classifier is not confident enough.


  \subsection{Scene Classifier} 

Additionally, we considered a scene classifier \cite{zhou2017places} to extract scene information from each image. We used a pre-trained scene classifier {\textit{Places365-ResNet}}\footnote{\href{http://places2.csail.mit.edu/}{http://places2.csail.mit.edu/}} to extract scene categories. 

According to the authors of \textit{Places365-ResNet} the network achieved good result in \textit{top-5 accuracy}, which make it ideal for multiple visual context extraction. The output from this classifier is a $365$ scene categories. Also, we consider a threshold to extract most likely classes in the images, and eliminate low confidence predictions.

\subsection{Semantic Similarity}
We aim to re-rank the baseline output using the visual context information, i.e. the semantic relation between the candidate words and the objects in the image.
We use a pre-trained visual classifier to detect objects-scenes in the image and devise a strategy to reward candidate words that are semantically related to them. As shown in the example of Figure \ref{fig:happy} the top position of the re-ranking yields \textit{quarters} as the most semantically related with the top position re-ranked object in the image \textit{parking}. 

Once the objects-scene in the image have been detected, we compute their semantic relatedness with the candidate words based on their word-embeddings \cite{Tomas:13}. Specifically, let us denote by $\vec{w}$ and $\vec{c}$ the word-embeddings of a candidate word $w$ and the most likely object $c$ detected in the image. We then compute their similarity using the cosine of the embeddings:
\begin{equation} 
\small sim(w, c) = \frac{\vec{w}\cdot \vec{c}}{|\vec{w}|\cdot|\vec{c}|}\;
\end{equation}

\section{Re-ranking Word Hypothesis }
\label{sec:reranking} 
In this section we describe the different re-rankers we devised for the list of candidate words produced by the baseline DNN of Sect.~\ref{sec:baselinesystems}.

\subsection{Unigram Language Model (ULM)}
 The first and simpler re-ranker we introduce is based on a word Unigram Language Model (ULM). The probabilities of the unigram model computed from  {\textit{Opensubtitles}}\footnote{\href{https://opensubtitles.org}{https://opensubtitles.org}} \cite{tiedemann2009news} and {\textit{Google book n-gram}}\footnote{\href{https://books.google.com/ngrams}{ https://books.google.com/ngrams}} 
 text corpora. The main goal of ULM is to increase the probability of the most common words proposed by the baseline. 
\begin{equation}
\small P_{ULM}(w)\;=\;\frac{count(w_j)}{\sum_{w\in C} count(w)} 
\end{equation}         

It is worth mentioning that the language model is very simple to build, train, and adapt to new domains, which opens the possibility of improving baseline performance for specific applications.

\subsection{Semantic Relatedness with Word Embedding (SWE)}
\label{sec:vci}
This re-ranker relies on the similarity between the candidate word and objects-scenes detected in the image. We compute (SWE) in the following steps: First, we use a threshold $\beta$ to eliminate lower probabilities from the visual classifier (objects, scenes). Secondly, we compute the similarity of each visual with the candidate word. Thirdly, we take the max-highest similarity score, most semantically related, to the candidate word $C_{max}$ as :        
\begin{equation}
    C_{max}= \argmax_{%
       \substack{%
         \phantom{\text{s.\,t.}}\, c_i \in Image \\  
         \phantom{\text{s.\,t.}}\, P(c_i) \geq \beta   
       }
     }
     sim(w,c_i) 
   \end{equation}
  
  Finally, following \cite{Sergey:03} with confirmation assumption $p(w|c) \geqslant p(w)$, we compute the conditional probability from similarity as:

\begin{equation}
\label{eq:sim}
\small P_{SWE}(w\vert c_{max})=P(w)^\alpha  \mathrm{~~~~~where~} \alpha=\left({\textstyle\frac{1-sim(w,c_{max})}{1+sim(w,c_{max})}}\right)^{1-P(c_{max})\;}
\end{equation}

where $P(w)$ is the probability of the word in general language (obtained from the unigram model), and $P(c_{max})$ is the probability of the most semantically related context object or places to the spotted text (obtained from the visual classifier). 

Note that  Equation \ref{eq:sim} already includes frequency information from the ULM, therefore it is taking into account not only the semantic relatedness, but also the word frequency information used in the ULM re-ranker above. Also, the ULM act alone in case there is no visual context information.

\subsection{Estimating Relatedness from Training Data Probabilities (TDP)}

A second possibility to compute semantic relatedness is to estimate it from training data. This should overcome the word embedding limitation when the candidate word and the image objects are not semantically related in general text, but are in the real world. For instance, as shown in the top-left example of Figure \ref{fig:moreexamples}, the sports TV channel \textit{kt} and the object \textit{racket} have no semantic relation according to the word embedding model, but they are found paired multiple times in the training dataset, which implies they do have a relation. 
For this, we use training data to estimate the conditional probability $P_{TDP}(w \vert c)$ of a word $w$ given that object $c$ appears in the image:
\begin{equation}
\small P_{TDP}(w\vert c)\;=\;\frac{count(w,c)}{count(c)}
\end{equation}
Where $count(w,c)$ is the number of training images where $w$ appears as the gold standard annotation for recognized text, and the object classifier detects object $c$ in the image. Similarly, $count(c)$ is the number of training images where the object classifier detects object $c$.

\subsection{Semantic Relatedness with Word Embedding (revisited) (TWE)}
This re-ranker builds upon a word embedding, as the SWE re-ranker above, but the embeddings are learnt from the training dataset (considering two-word ``sentences'': the target word and the object in the image). The embeddings can be computed from scratch, using only the training dataset information (TWE) or initialized with a general embeddings model that is then biased using the training data (TWE*). 

In this case, we convert the similarity produced by the embeddings to probabilities using:
\begin{equation}
\label{eq:tanh}
\small  \small P_{TWE}(w\vert c)=\frac{\tanh(sim(w,c))+1}{2 P(c)}
\end{equation}
Note that this re-ranker does not take into account word frequency information as in the case of the SWE re-ranker.

\section{Combining Re-rankers}
\label{sec:combining}

Our re-ranking approach consists in taking the softmax probabilities computed by the baseline DNN and combine them with the probabilities produced by the re-ranker methods described in Section~\ref{sec:reranking}. We combine them by simple multiplication, which allows us to combine any number of re-rankers in cascade. 
We evaluated the following combinations:

\begin{enumerate}
\item The baseline output is re-ranked by the unigram language model:
\begin{equation}
\small P_1(w)=P_{BL}(w)\times P_{ULM}(w)
\label{eq:ULM}
\end{equation}
\item The baseline output is re-ranked by the general word-embedding model (SWE). Note that this reranker also includes the ULM information.
\begin{equation}
\small P_2(w,c)=P_{BL}(w)\times P_{SWE}(w\vert c)
\label{eq:SWE}
\end{equation}
\item  The baseline output is re-ranked by the relatedness estimated from the training dataset as conditional probabilities (TDP).  
\begin{equation}
\small P_3(w,c)=P_{BL}(w)\times P_{TDP}(w\vert c)
\label{eq:TDP}
\end{equation}

\item  The baseline output is re-ranked by the word-embedding model trained entirely on training data (TWE) or a general model tuned using the training data (TWE*):
\begin{equation}
\small P_4(w,c)=P_{BL}(w)\times P_{TWE}(w\vert c)
\label{eq:TWE}
\end{equation}

\item We also apply SWE and TDP re-rankers combined: 
\begin{equation} \label{eq:SWE+TDP}
\small P_5(w,c)=P_{BL}(w)\times P_{SWE}(w\vert c)\times P_{TDP}(w\vert c)
\end{equation}

\item The combination of TDP and TWE:
\begin{equation} \label{eq:TDP+TWE}
\small P_6(w,c)=P_{BL}(w)\times P_{TDP}(w\vert c)\times P_{TWE}(w\vert c)
\end{equation}

\item Finally, we combine all re-rankers together:
\begin{equation} \label{eq:all}
\small 
\begin{split}
\small P_7(w,c)\;=\;P_{BL}(w)\;\times\;P_{SWE}(w\vert c)\;  
\small \times P_{TDP}(w\vert c)\times P_{TWE}(w\vert c)
\end{split}
\end{equation}

\end{enumerate}

\section{Experiments and Results}
\label{sec:experiments}

In this section we evaluate the performance of the proposed approaches in the 
{\bf ICDAR-2017-Task3 (end-to-end)} \cite{Andreas:16} dataset.
This dataset is based on Microsoft COCO \cite{Tsung-Yi:14} (Common Objects in Context), which consists of 63,686 images, and 173,589 text instances (annotations of the images). COCO-Text was not collected with text recognition in mind, therefore, not all images contain textual annotations. The \textit{ICDAR-2017 Task3} aims for end-to-end text spotting (i.e. both detection and recognition). Thus, this dataset includes whole images, and the texts in them may appear rotated, distorted, or partially occluded. Since we focus only on text recognition, we use the ground truth detection as a golden detector to extract the bounding boxes from the full image. The dataset consists of 43,686 full images with 145,859 text instances, and for training 10,000 images and 27,550 instances for validation. We evaluate our approach on a subset \footnote{\href{https://github.com/ahmedssabir/dataset/}{https://github.com/ahmedssabir/dataset/}} of the validation containing 10,000 images with associated bounding boxes.
\subsection{Preliminaries} 
For evaluation, we used a more restrictive protocol than the standard proposed by \cite{Kai:11} and adopted in most state-of-the-art benchmarks, which does not consider words with less than three characters or with non-alphanumerical characters. This protocol was introduced to overcome the false positives on short words that most current state-of-the-art struggle with, including our Baselines. However, we overcome this limitation by introducing the language model re-ranker. Thus, we consider all cases in the dataset, and words with less than three characters are also evaluated. 

In all cases, we use two pre-trained deep models, CNN \cite{Max:16} and LSTM \cite{Suman:17} as a baseline (BL) to extract the initial list of word hypotheses. Since these BLs need to be fed with the cropped words, when evaluating on the ICDAR-2017-Task3 dataset we will use the ground truth bounding boxes of the words.

\subsection{Experiments with Language Model}

As a proof of concept, we trained our unigram language model on two different copora. The first ULM was trained on {\textit{Opensubtitles}\cmt{\footnote{\href{https://www.opensubtitles.org}{\textit{https://www.opensubtitles.org}}}}, a large database of subtitles for movies containing around 3 million word types, including numbers and other alphanumeric combinations that make it well suited for our task. Secondly, we trained another model with {\textit{Google book n-gram}}\cmt{\footnote{\href{https://books.google.com/ngrams}{\textit{https://books.google.com/ngrams}}}}, that contains 5 million word types from American-British literature books. However, since the test dataset contains numbers, the accuracy was lower than that obtained using the \textit{Opensubtitles} corpus. We also evaluate  a model trained on the union of both corpora, that contains around 7 million word types.

In this experiment, we extract the $k=2,\ldots,9$ most likely words --and their probabilities-- from the baselines. Although the sequential nature of the LSTM baseline captures a character-based language model, our post-process uses word-level probabilities to re-rank the word as a whole. Note that since our baselines work on cropped words, we do not evaluate the whole end-to-end but only the influence of adding external knowledge.  

The first baseline is a CNN \cite{Max:16} with fixed-lexicon recognition, which is not able to recognize any word outside its dictionary. The results are reported in Table \ref{table_1}. We present three different accuracy metrics: 1) \textit{full} columns correspond to the accuracy on the whole dataset, while 2) \textit{dictionary} columns correspond to the accuracy over the solvable cases (i.e. those where the target word is among the 90K-words of the CNN dictionary, which correspond to 43.3\% of the whole dataset), and finally 3) \textit{list} shows the accuracy over the cases where the right word was in the $k$-best list output by the baseline. 
We also provide the results using different numbers of $k$-best candidates. Table \ref{table_1} top  row shows the performance of the CNN baseline, and the second row reports the influence of the ULM. The best result is obtained with $k=3$, which improved the baseline model in 0.9\%, up to 22\% \textit{full}, and 2.7\%, up to 61.3\%  \textit{dictionary}.

The second  baseline we consider is an LSTM \cite{Suman:17} with visual soft-attention mechanism, performing unconstrained text recognition without relying on a lexicon. The first row in Table \ref{table_2} reports the LSTM baseline result on this dataset, and the second row shows the results after the ULM re-ranking. The best results are obtained by considering $k=3$ which improves the baseline in 0.7\%, from 18.72\% to 19.42\%.

In summary, the lexicon-based baseline CNN performs better than the unconstrained approach LSTM, since the character sequences prediction generation that may lead up to random words, which the ULM may be unable to re-rank.
\begin{table}[t] 
\centering 

\caption{Results of re-ranking the $k$-best ($k=2\ldots9$) hypotheses of the CNN baseline on ICDAR-2017-Task3 dataset (\%)} 
\begin{threeparttable}
\renewcommand{\arraystretch}{1.3} 
\centering 
\small
\begin{tabular}{|l|lll|lll|lll|lll|}

\hline 

\textbf{Model}  & \multicolumn{3}{c|}{\textbf{ $k= 2$}}   & \multicolumn{3}{c|}{\textbf{ $k= 3$}}   & \multicolumn{3}{c|}{\textbf{$k= 5$}} & \multicolumn{3}{c|}{\textbf{$k= 9$}}              \\

         &  full  &   dict   &   list &  full    &  dict  & list &   full  &  dict & list & full&  dict & list\\    
\hline\hline
 \textit{CNN baseline$_1$} & \multicolumn{12}{c|}{\textit{full: 21.1 \space dictionary: 58.6}} \\
\hline
 CNN+ULM$_{\text{7M}}$   &    21.8      &    60.6   & 90.0 &   22.0       &     61.3   &   84.2   &     21.6       &     60.1  &  77.7  &   21.0     &   58.5  &  68.7   \\ \hline
CNN+SWE$_{object}$             &     22.3     &    62.1   &  92.3  &   22.6       &     63.0    & 86.5    &     22.8       &     63.4   &  81.9  &   22.6    &   62.9  & 73.9 \\ 
CNN+SWE$_{place}$             &     22.1     &    61.4    &  91.2 &  22.5        &     62.5     &    85.8&     22.6       &     62.6    &  80.8 &   22.6    &   62.8  &  73.8\\
\hline
CNN+TDP                 &   22.2       &    61.7     & 91.6 &   22.7      &     63.3    &   86.9  &     22.7       &     63.2 &   81.6   &  22.6     &   62.8 &         73.8    \\
\hline
CNN+SWE$_{object}$+TDP          &   22.4       &    62.2    &   92.4 &  22.9       &  63.6          & 87.4 &     \textbf{23.0}      &     \textbf{64.0}     & \textbf{82.6} &  22.9      &  63.7     &74.8 \\ 
 CNN+SWE$_{place}$+TDP         &   22.1      &    61.6    &   91.5 &  22.6       &  62.7         &  86.1  &     22.8       &     63.4     & 81.9 &  22.8      &  63.4    & 74.5\\ 

\hline
CNN+TWE                &   22.3       &    61.9     & 92.0   &  22.6       &  62.9      &    86.4  &     22.6       &     62.8 &   81.1   &  22.7      &   63.0 &         74.0         \\ 
\hline
CNN+TDP+TWE*           & 22.3         &   62.1      &  92.3  &  22.8    &     63.4    &    87.0    &      22.9    &     63.8  &   82.4     &  \textbf{23.0}          &    \textbf{64.0} &    \textbf{75.2}   \\ 
\hline
CNN+All$_{object}$    & 22.3         & 62.1     & 92.3      & 22.7        & 63.2   &    86.7      &  22.9          & 63.6  &   82.1      &  22.7      & 63.3 & 74.3 \\
CNN+All$_{place}$   & 22.2       & 61.8       & 91.9     & 22.7        & 63.1      &  86.6     &  22.8          & 63.4     &   81.9   &  22.6      & 63.0 &  74.0 \\  
\hline

\end{tabular} 

\end{threeparttable}
\label{table_1} 
\end{table}  
\subsection{Experiments with Visual Context Information}
\label{experiment2}
The main contribution of this paper consists in  re-ranking the $k$ most likely hypotheses candidate word using the visual context information. Thus, we use ICDAR-2017-Task3 dataset to evaluate our approach, re-ranking the baseline output using the semantic relation between the spotted text in the image and its visual context.  
 As in the language model experiment, we used ground-truth bounding boxes as input to the BL. However, in this case, the whole image is used as input to the visual classifier.

In order to extract the visual context information we considered two different pre-trained state-of-the-art visual classifiers: object and scene classifiers. For image classification we rely on three pre-traind network:   
ResNet \cite{Kaiming:16}, GoogLeNet \cite{Christian:15} and Inception-ResNet-v2 \cite{szegedy2017inception}, all of them able to detect pre-defined list of 1,000 object classes. However, for testing we considered only Inception-ResNet-v2 due to better \textit{top-5 accuracy}. For scene classification we use places classifier \textit{Place365-ResNet152} \cite{zhou2017places} that able to detect 365 scene categories. 

Although the visual classifiers use a softmax to produces only one probable object hypotheses per image, we use \textit{threshold} to extract a number of object-scene hypotheses, and  eliminate  low-confidence results. Then, we compute the semantic relatedness for each object-scene hypotheses with the spotted text. Finally, we take the most related visual context.

\begin{table}[t] 
\small 
\centering 
\caption{Results of re-ranking the $k$-best ($k=2\ldots9$) hypotheses of the LSTM baseline on ICDAR-2017-Task3 dataset (\%)} 
\begin{threeparttable}
\renewcommand{\arraystretch}{1.3} 
\centering 

\begin{tabular}{|l|ll|ll|ll|ll|}

\hline 

\textbf{Model}  & \multicolumn{2}{c|}{\textbf{ $k= 2$}}   & \multicolumn{2}{c|}{\textbf{ $k= 3$}}   & \multicolumn{2}{c|}{\textbf{$k= 5$}} & \multicolumn{2}{c|}{\textbf{$k= 9$}}              \\

         &  full  &    list &   full     & list &   full  &   list & full&   list\\    
\hline
\textit{LSTM baseline$_2$}  & \multicolumn{8}{c|}{\textit{18.7}} \\
 \hline
 LSTM+ULM$_{\text{7M}}$         &   19.3            &    79.7   & 19.4                    &   74.2  &     19.1     &68.3 &    18.7    &        60.5       \\ 
 \hline
LSTM+SWE$_{object}$                      &   19.3        &  80.0     &  19.8                 &   75.5   &      20.0       &   71.8     &   20.1       &   65.8       \\ 
LSTM+SWE$_{place}$                     &   19.3         &  79.8    &  19.7                  &     75.3 &      20.1        &     72.4   &   20.0         &    65.3    \\
 \hline
LSTM+TDP                        &   19.0           &     78.7 &  19.3                  &    73.7     &     19.5       &   70.1   &   20.0        &      65.4       \\ 
\hline 
LSTM+SWE$_{object}$+TDP                 &   19.4         &    80.5   &  20.0                    &    76.4  &     20.3        &   73.1   &  20.6          &   67.2        \\ 
LSTM+SWE$_{place}$+TDP                &   19.4       &      80.4   &  19.9                     &    75.9     &     20.3        &    72.9  &  20.4          &       66.6    \\ 
\hline 
LSTM+TWE                       &   19.5        &     80.6  &   20.0                  &    76.2    &     20.1      &     72.4  & 20.3        & 66.4\\
\hline
LSTM+TDP+TWE* 	               &   19.5        &      80.8   &  20.0                	 &    76.5   &   20.3       	&  73.1    &  \textbf{20.8}    &     \textbf{68.0}          \\ 
\hline 
LSTM+All$_{object}$         &   19.4      &     80.2     &    19.8              &       75.6          &   20.3              &  72.9 &   20.4          &    66.8            \\ 
LSTM+All$_{place}$      &   19.4        &    80.1    &    20.0                  &     76.2     &   20.3       &     72.8       &  20.3              &       66.5         \\

\hline
\end{tabular} 

\end{threeparttable}
\label{table_2} 
\end{table}

In this experiment we re-rank the baseline $k$-best hypotheses based on their relatedness with the objects in the image. We try two approaches for that: 1) semantic similarity computed using word embeddings \cite{Tomas:13} and 2) correlation based on co-ocurrence of text and image object in the training data. 

First, we re-rank the words based on their word embedding: semantic relatedness with multiple visual context from general text : 1) object  (SWE$_{object}$) and 2) scene (SWE$_{place}$). For instance, the top-right example in Figure~\ref{fig:moreexamples}   shows that the strong semantic similarity between scene information \textit{parking} and \textit{pay} re-ranked that word from 3rd to 1st position. We tested three pre-trained models trained on general text as baseline 1) word2vec model with 100 billion tokens 2) glove model with 840 billion tokens \cite{Jeffrey:14} and 3) fastText with 600 billion tokens \cite{bojanowski2017enriching}. However, we adopt glove as baseline, due to a better similarity score.

Secondly, we use the training data to compute the conditional probabilities between text image and object in the image happen together (TDP). We also combined both relatedness measures as described in Equation~\ref{eq:SWE+TDP}, obtaining a higher accuracy improvement on both baselines, as can be seen in Table \ref{table_1} and \ref{table_2}, (SWE+TDP) boosted the accuracy for both baseline. The LSTM accuracy improved up to 1.9\% . In other hand, the CNN, with 90k fixed lexicon, accuracy is boosted up to 1.9\% on \textit {full} dataset and 5.4\% \textit{dictionary}. For example, as shown in Figure~\ref{fig:moreexamples} top-left example, text image \textit{kt} (sport channel) happens often with visual context \textit{racket}, something that can not be captured by general word embedding models. Also, scene classifier SWE$_{place}$+TDP boost the baseline 1.7\% \textit{full} and 4.8\% \textit{dictionary}. The scene classifier SWE$_{place}$ perform better than the object classifier in instance outdoor. For instance, the spotted text in a signboard \textit{way} is more semantically related with \textit{downtown} than a man holding \textit{an umbrella} in the image.   

Finally, we trained a word embedding model using the training dataset (TWE). Due to the dataset is too small, we train skip-gram model with one window, and without any word filtering. In addition, we initialized the model weight with the baseline (SWE) that trained on general text, we call it TWE*. The result is 300-dimension vector for about 10K words. Also, we initialized the weight randomly but when we combined the re-rankers the pre-trained initialized model is slightly better. The result in both Table \ref{table_1} and \ref{table_2} button two rows shows that (TWE) outperform the accuracy of SWE model that trained on general text.   

The result in Table \ref{table_1} CNN shows that the combination model TDP+TWE also significantly boost the accuracy up to 5.4\% \textit{dictionary} and 1.9\% \textit{all}. Also, in Table \ref{table_2}, the second baseline LSTM accuracy boosted up to 2.1\%. Not to mention that TDP+TWE model only rely on the visual context information, computed by Equation \ref{eq:tanh}.     
\begin{table}[t]
\begin{center}
\caption{Examples of $P(word|object)$ for each re-ranker. TDP and TWE capture relevant information to improve the baseline for pairs word-object/scene that appear in the training dataset. The TPD overcome word-embedding limitation in samples happen in training datasets.}

\begin{threeparttable}

\begin{tabular}{|l|l|l|l|l|l|}
\hline 
 Word &  Visual    &  SWE &  TDP &  TWE &  TWE* \\
\hline \hline
delta     & airliner &  0.0028  & \textbf{0.0398}  & 0.0003  &  0.00029     \\
kt         & racket  & 0.0004 &   \textbf{0.0187} &   0.0002 & 0.00006 \\ 
plate     & moving   &  0.0129  & 0.00050  & \textbf{0.326}  &  0.00098       \\
way       & street   & 0.1740    & 0.02165  &   \textbf{0.177}       &  0.17493  \\ 
\hline
\end{tabular}

\end{threeparttable}
\end{center}

 \label{table_3} 
 \end{table}

\section{Discussion}  
\label{sec:discussion} 
The visual context information re-ranks potential candidate words based on the semantic relatedness with its visual information (SWE). However, there are some cases when there is no direct semantic correlation between the visual context and the potential word.  Thus we proposed TDP to address this limitation by learning correlations from the training dataset. However, there are still  cases unseen in the training dataset, for instance, as shown in Figure~\ref{fig:moreexamples} bottom-left text image \textit{copyrighting} and its visual context \textit{ski slop, snowfield} have neither semantic correlation nor were seen in the training dataset. There are also cases where is no relation at all, as in Figure \ref{fig:moreexamples} the brand name \textit{zara} and the visual context \textit{crosswalk} or \textit{plaza}. 

The results we have presented  show that our approach is a simple way to boost accuracy of text recognition deep learning models, or to adapt them to particular domains,  with a very low re-training/tuning cost. The proposed post-processing approach can be used as a drop-in complement for any text-spotting algorithm (either deep-learning based or not) that outputs a ranking of word hypotheses.
In addition, our approach overcomes some of the limitations that current state-of-the-art deep model struggle to solve in complex background text spotting scenarios, such as short words. 

One limitation of this approach is that when the language model re-ranker is strong, the visual context re-ranker is unable to  re-rank the correct candidate word. For instance,  the word \textit{ohh} has a large frequency count in general text. This problem can be tackled by adjusting the weight of uncommon short words in the language model. 
\begin{figure*}[t]
\centering 

\includegraphics[width=4.6in]{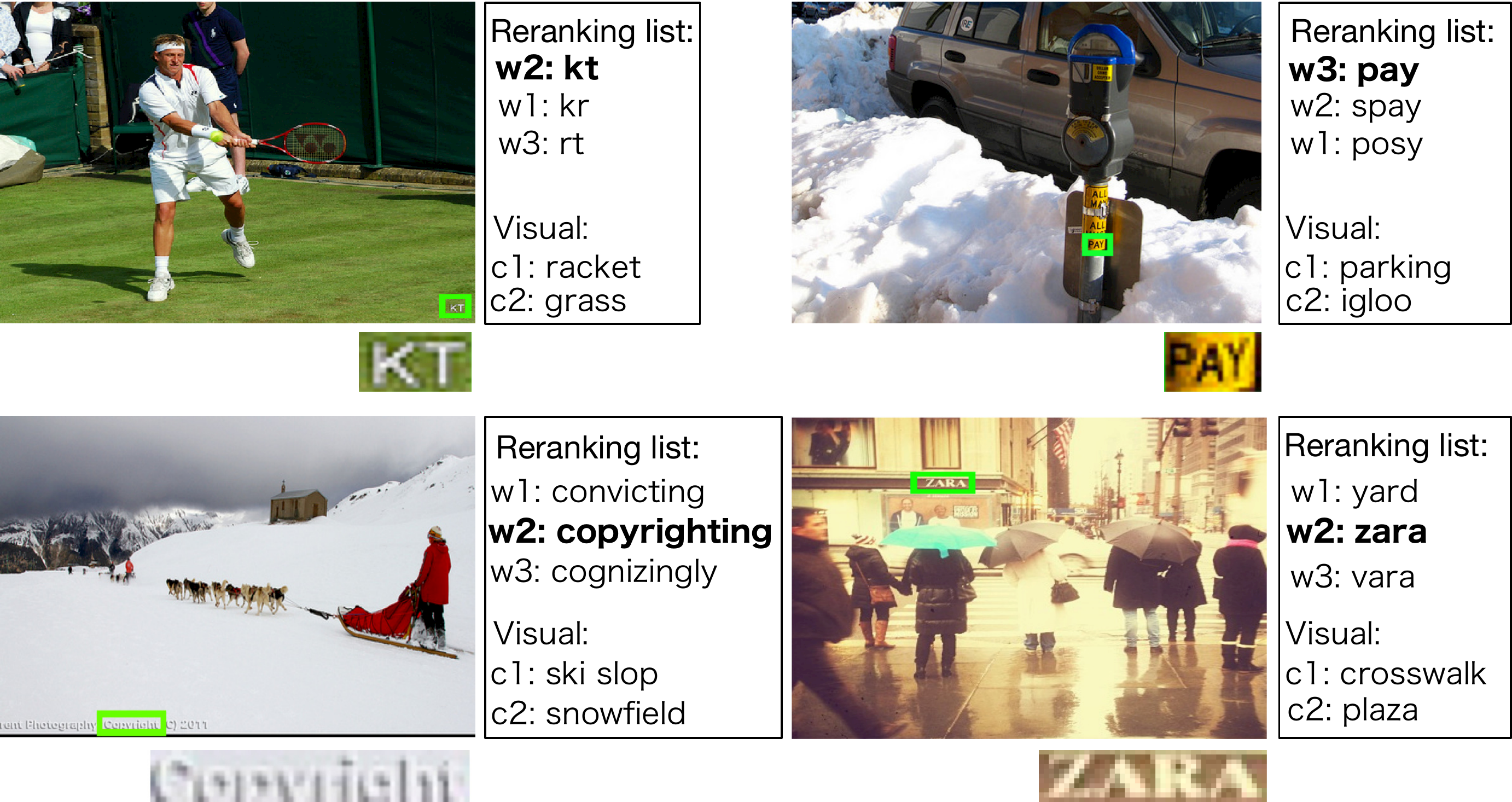}
\caption{Some examples of visual context re-ranker. The top-two examples are successful results of the visual context re-ranker. The top-left example is a re-ranking result based on the relation between text and its visuals happen together in the training dataset. The top-right example is a re-ranking result based on semantic relatedness between the text image and its visual. The two cases in the bottom are examples of words either have no semantic correlation with the visual or exist in the training dataset. Not to mention that the top ranking visual $c_1$ is the most semantically related visual context to the spotted text. (Bold font words indicate the ground truth)}

\label{fig:moreexamples} 
\end{figure*}

\section{Conclusion}
\label{sec:conclusion}
In this paper we have proposed a simple post-processing approach, a hypothesis re-ranker based on visual context information, to improve the accuracy of any pre-trained text spotting system. We  also show that by integrating a language model re-ranker as a prior to the visual re-ranker, the performance of the visual context re-ranker can be improved. We have shown that the accuracy of two state-of-the-art deep network architectures, a lexicon-based and lexicon-free recognition, can be boosted up to 2 percentage-points on standard benchmarks. In the future work, we plan to explore end-to-end based fusion schemes that can automatically discover more proper priors in one shot deep model fusion architecture. 


\bibliographystyle{splncs}
\bibliography{egbib}

\end{document}